\documentclass{article}

     \usepackage[final]{neurips_2019}

\usepackage[utf8]{inputenc} 
\usepackage[T1]{fontenc}    
\usepackage{hyperref}       
\usepackage{url}            
\usepackage{booktabs}       
\usepackage{amsfonts}       
\usepackage{nicefrac}       
\usepackage{microtype}      

\usepackage{times}
\usepackage{soul}
\usepackage{latexsym}

\usepackage{enumerate}
\usepackage{graphicx}
\usepackage{subfigure}
\usepackage{multirow}

\usepackage{framed,multirow}
\usepackage{times}
\usepackage{graphicx} 
\usepackage{subfigure}
\usepackage{amssymb}
\usepackage{latexsym}

\usepackage{algorithm}
\usepackage{algorithmic}
\usepackage{hyperref}

\usepackage{url}
\usepackage{xcolor}
\usepackage{amsmath}
\usepackage{amsxtra}
\usepackage{braket}
\usepackage{array}
\usepackage{color}
\usepackage{url}
\usepackage{placeins}
\usepackage{threeparttable}

\usepackage{times}
\usepackage{balance}
\usepackage{enumerate}
\usepackage{multirow}
\usepackage{pifont}
\usepackage{marvosym}
\usepackage{ifsym}

\title{Fused Lasso for Feature Selection using Structural Information}

\author{%
  Lu Bai${}^{1,3}$, Lixin Cui${}^{1}$, Yue Wang${}^{1}$, Philip S. Yu${}^{2}$, Edwin R. Hancock${}^{3}$\\
  ${}^{1}$Department of Computer Science, Central University of Finance and Economics, Beijing, China\\
  \texttt{\{bailucs, cuilixin, wangyuecs\}@cufe.edu.cn} \\
  ${}^{2}$Department of Computer Science, University of Illinois at Chicago, Chicago, USA\\
  \texttt{psyu@uic.edu} \\
  ${}^{3}$Department of Computer Science, University of York, York, UK\\
  \texttt{edwin.hancock@york.ac.uk}
}
\begin{document}

\maketitle

\begin{abstract}
Feature selection has been proven a powerful preprocessing step for high-dimensional data analysis. However, most state-of-the-art methods tend to overlook the structural correlation information between pairwise samples, which may encapsulate useful information for refining the performance of feature selection. Moreover, they usually consider candidate feature relevancy equivalent to selected feature relevancy, and some less relevant features may be misinterpreted as salient features. To overcome these issues, we propose a new feature selection method using structural correlation between pairwise samples. Our idea is based on converting the original vectorial features into structure-based feature graph representations to incorporate structural relationship between samples, and defining a new evaluation measure to compute the joint significance of pairwise feature combinations in relation to the target feature graph. Furthermore, we formulate the corresponding feature subset selection problem into a least square regression model associated with a fused lasso regularizer to simultaneously maximize the joint relevancy and minimize the redundancy of the selected features. To effectively solve the optimization problem, an iterative algorithm is developed to identify the most discriminative features. Experiments demonstrate the effectiveness of the proposed approach.
\end{abstract}

\section{Introduction}\label{s1}
High-dimensional data are ubiquitous in many machine learning applications. Such high-dimensional data poses significant challenges for classifications, since they not only demand expensive computational complexity but also degrade the generalization ability of the learning algorithm~\cite{DBLP:books/lib/HastieTF09,DBLP:journals/pr/GaoHZ18}. One effective way to deal with this problem is feature selection~\cite{DBLP:conf/ijcai/ZhengZZZ18}. By discarding irrelevant and redundant features, feature selection directly identifies a subset of the most discriminative features from the original feature space so that the classification accuracy and interpretability of the learning algorithm can be improved~\cite{DBLP:journals/kbs/WangSHQQ16}. Depending on how they utilize the learning algorithm in the search process, feature selection methods can be partitioned into a) filter methods~\cite{DBLP:journals/tnn/DitzlerPR18}, b) wrapper methods~\cite{DBLP:journals/kbs/WangACLA15} and c) embedded methods~\cite{DBLP:journals/pr/GaoHZ18}. The filter methods have preferable generalization ability and high computation efficiency~\cite{DBLP:journals/ijon/QianS15}, thus they are usually preferred in many real-world applications.

In the literature, many efficient filter methods have been proposed based on various criteria used for evaluating feature subsets, such as consistency~\cite{DBLP:journals/ai/DashL03}, correlation~\cite{DBLP:conf/icml/Hall00} and mutual information (MI) ~\cite{DBLP:journals/pr/HermanZ0YC13}, etc. Among these, MI measures are considered to be most effective as they are able to measure the nonlinear relationships between features and the target variables~\cite{DBLP:journals/pr/HermanZ0YC13}. Existing MI-based methods mostly concentrate on maximizing dependency and relevancy or minimizing redundancy. Representative examples include 1) the mutual information-based feature selection (MIFS)~\cite{DBLP:journals/tnn/Battiti94}, 2) the maximum-relevance minimum-redundancy criterion (MRMR)~\cite{DBLP:journals/pami/PengLD05}, 3) the joint mutual information maximisation criterion (JMIM)~\cite{DBLP:journals/eswa/BennasarHS15}, etc.

Although efficient, previous vectorial-based information theoretic feature selection methods may yield suboptimal subsets due to the following reasons. First, they usually overlook the structural correlation information between pairwise samples, which may encapsulate useful information for refining the performance of feature selection. More specifically, denote a dataset consisting of $N$ features and $M$ samples as $\mathcal{X}=\{\mathbf{f_{1}},\ldots,\mathbf{f_{i}},\ldots,\mathbf{f_{N}}\}$, where $\mathbf{f_{i}}$ represents each feature and $\mathbf{f_{i}}=(f_{i1},\ldots,f_{ia},\ldots,f_{ib},\ldots,f_{iM})^T$ is the corresponding vectorial representation for each feature $\mathbf{f_{i}}$. By calculating various mutual information measures between feature vectors, one can select an optimal feature subset. However, such calculation can not incorporate the relationship between pairwise samples $f_{ia}$ and $f_{ib}$ in $\mathbf{f_{i}}$ into the feature selection process and may lead to significant information loss. For example, given three marbles denoted as M1 (Sphere, Red Colored), M2 (Triangle, Pink Colored), and M3 (Ellipsoid, Blue Colored). If we classify them according to color, M1 and M2 will be in the same class. On the other hand, if we classify them in terms of shape, M1 and M3 will be in the same class. This example indicates that sample relationships associated with various features are different and are important to evaluate the effectiveness of features. Second, they usually consider individual feature relevancy equivalent to selected feature relevancy, i.e., feature relevancy is usually defined as the mutual information between a candidate feature and the target, without considering joint relevancy of pairwise features. Therefore, some less relevant features may be misinterpreted as salient features.

\begin{figure}
\centering
{\includegraphics[width=0.99\linewidth]{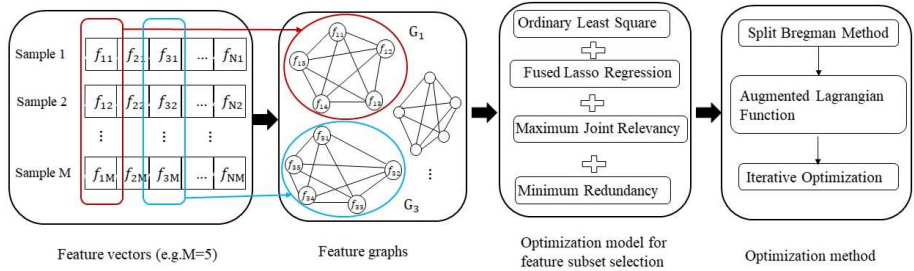}}
\vspace{-30pt}
\caption{Framework of the proposed feature selection method} \label{embeddingsB}
\vspace{-20pt}
\end{figure}

In feature selection, the appealing characteristics of graph representations have facilitated the development of some pioneering works to tackle these issues. For instance, \cite{DBLP:journals/corr/abs-1809-02860} have introduced a novel feature selection approach based on graph representations. Our approach significantly differs from \cite{DBLP:journals/corr/abs-1809-02860} and the major contributions of this work are highlighted as follows. Besides, the framework of the proposed feature selection approach is presented in Figure~\ref{embeddingsB}.

\textbf{First}, unlike \cite{DBLP:journals/corr/abs-1809-02860}, we propose a kernel-based modelling method to convert each original vectorial feature into a structure-based feature graph representation, for the objective of encapsulating structural correlation information between pairwise samples. Furthermore, a new structural information theoretic measure associated with the feature graph representations is developed to simultaneously maximize joint relevancy of different pairwise feature combinations in relation to the discrete targets and minimize redundancy among selected features.

\textbf{Second}, with the proposed structural information theoretic measure to hand, we compute an interaction matrix to characterize the structural informative relationship between pairwise feature combinations in relation to the discrete target. Moreover, we formulate the corresponding feature subset selection problem into the framework of a least square regression model associated with a fused lasso regularizer. The reasons of using fused lasso rather than elastic net are as follows. First, when the number of features is larger than the sample size, elastic net may be less efficient than other regularization terms such as fused lasso. Second, fused lasso can ensure sparsity in both the coefficients and differences of successive coefficients.

\textbf{Third}, because of the nonseparability and nonsmoothness of the fused lasso regularizer in its objective function, an efficient iterative algorithm is developed to locate the optimal solutions to the proposed feature subset selection problem. The experiments verify the effectiveness of the proposed feature selection approach.


\section{Preliminary Concepts}\label{s2}
In this section, we first illustrate the construction of the feature graph which incorporates structural correlation information of pairwise feature samples. Then we review the preliminaries of Jensen-Shannon divergence, which is utilized to compute the similarity between feature graph structures.
\subsection{Kernel-based Feature Graph Modelling}
In this subsection, we illustrate how to convert the original vectorial features into structure-based feature graphs, in terms of a kernel-based similarity measure. The reason of representing each original feature as a graph structure is that graph-based representation can capture richer global topological information than vectors. Thus, the pairwise sample relationships of each original feature vector can be incorporated into the selection process of the most discriminative features to reduce information loss. Let $\mathcal{X}=\{\mathbf{f_{1}},\ldots,\mathbf{f_{i}},\ldots,\mathbf{f_{N}}\}\in R^{M\times N}$ be a dataset of $N$ features and $M$ samples. As shown in Figure~\ref{embeddingsB}, we transform each original vectorial feature $\mathbf{f_{i}}=(f_{i1},\ldots,f_{ia},\ldots,f_{ib},\ldots,f_{iM})^T$ into a feature graph structure $\mathbf{G_{i}}(V_i,E_i)$, where each vertex $v_{ia}\in V_i$ represents the $a$-th sample $f_{ia}$ and each weighted edge $(v_{ia},v_{ib})\in E_i$ represents the relationship between the $a$-th and $b$-th samples.
For example, assume $\mathbf{G_{1}}$ and $\mathbf{G_{3}}$ are the graph-based representations of features $\mathbf{f_{1}}$ and $\mathbf{f_{3}}$. Given a pair of samples $\mathbf{Sample}$ $1$ and $\mathbf{Sample}$ $2$, the corresponding sample correlations in relation to features $\mathbf{f_{1}}$ and $\mathbf{f_{3}}$, denoted as edges $(f_{11},f_{12})\in E_1$ and $(f_{31},f_{32})\in E_3$, are quite different. This indicates that sample correlation can be incorporated into the feature selection process using graph-based representations.

Moreover, we also need to construct a graph structure for the target feature $\mathbf{Y}$. For classification problems, $\mathbf{Y}$ are the class labels and usually take the discrete class values $c\in\{1,2,\ldots,C\}$. For such case, we first compute the continuous value based target feature for each feature $\mathbf{f_{i}}$ as $\mathbf{\hat{f}_{i}}=(\hat{f}_{i1},\ldots,\hat{f}_{ia},\ldots,\hat{f}_{ib},\ldots,\hat{f}_{iM})^T$, where each element $\hat{f}_{ia}$ corresponds to the $a$-th sample. When the element ${f}_{ia}$ of $\mathbf{{f}_{i}}$ belongs to the $c$-th class, the value of $\hat{f}_{ia}$ is the mean value $\mu_{ia}$ of all samples in $\mathbf{{f}_{i}}$ from the same class $c$. Similar to the process of converting each original feature $\mathbf{f_{i}}$ into the feature graph, we construct the resulting target feature graph representation for each feature $\mathbf{f_{i}}$ associated with its continuous value based target feature $\mathbf{\hat{f}_{i}}$ as $\mathbf{\hat{G}_{i}}(\hat{V}_{i},\hat{E}_{i})$, where each vertex $\hat{v}_{ia}$ represents the $a$-th
sample of $\mathbf{\hat{f}_{i}}$ (i.e., the $a$-th sample of $\mathbf{Y}$ in terms of $\mathbf{\hat{f}_{i}}$), and $(v_{ia},v_{ib})\in E_i$ represents the relationship between the $a$-th and $b$-th samples of $\mathbf{f_{i}}$ (i.e., the structural relationship between the $a$-th and $b$-th samples of $\mathbf{Y}$ in terms of $\mathbf{\hat{f}_{i}}$). To compute the relationship between pairwise feature samples, \cite{DBLP:journals/corr/abs-1809-02860} have employed the Euclidean distance as the measure to construct both the feature graph $\mathbf{{G}_{i}}({V}_{i},{E}_{i})$ and the target feature graph $\mathbf{\hat{G}_{i}}(\hat{V}_{i},\hat{E}_{i})$. However, the characteristics of these graph structures may be overemphasized and dominated by the large distance value.

To overcome the aforementioned problem, we further propose a new kernel-based similarity measure associated with the original Euclidean distance to construct the (target) feature graph structures. Specifically, for the feature graph $\mathbf{{G}_{i}}({V}_{i},{E}_{i})$ of $\mathbf{{f}_{i}}$ and its associated Euclidean distance based adjacency matrix $A$, each row (column) of $A$ can be seen as the distance based embedding vector for each sample of $\mathbf{{f}_{i}}$. Assume $A_{a,:}$ and $A_{b,:}$ denote the embedding vectors of the $a$-th and $b$-th samples respectively. The relationship between these two samples can be computed as their normalized kernel value associated with dot product
\begin{equation}
K_{a,b}=\frac{< A_{a,:},A_{b,:}>}{\sqrt{< A_{a,:},A_{a,:}> < A_{b,:},A_{b,:}> }},
\end{equation}
where $< \cdot,\cdot >$ is the dot product. We utilize the kernel matrix to replace the original Euclidean distance matrix as the adjacency matrix of $\mathbf{{G}_{i}}({V}_{i},{E}_{i})$, and the relationships between the samples of $\mathbf{{f}_{i}}$ are all bounded between $0$ and $1$. For the target feature graph $\mathbf{\hat{G}_{i}}(\hat{V}_{i},\hat{E}_{i})$, we also compute its adjacency matrix using the same procedure. The kernel-based similarity measure not only overcomes the shortcoming of graph characteristics domination by the large Euclidean distance value between pairwise feature samples, but also encapsulates high-order relationship between feature samples. This is because the kernel-based relationship between each pair of samples associated with their distance based embedding vector encapsulates the distance information between each feature sample and the remaining feature samples. Finally, the kernel-based relationship can also represent the original vectorial features in a high-dimensional Hilbert space, and thus reflect richer structural characteristics.
\subsection{The JSD for Multiple Probability Distributions}
In Statistics and Information Theory, an extensively used measure of dissimilarity between probability distributions is the Jensen–Shannon divergence (JSD)~\cite{DBLP:journals/tit/Lin91}. JSD has been successful in a wide range of applications, including analysis of symbolic sequences and segmentation of digital images. In~\cite{DBLP:conf/pkdd/Bai0BH14}, the JSD has been adopted to measure similarity between graphs associated with their probability distributions. Moreover, \cite{DBLP:journals/corr/abs-1809-02860} have utilized the JSD to compute the similarity between an individual feature graph in relation to its target feature graph. Unlike the previous works that focus on the JSD measure between pairwise graph structures, our major concern is the similarity between multiple graphs. Specifically, the JSD measure can be used to compare $n$ ($n\geq2$) probability distributions
\begin{equation}\label{Eq:CJSD}
{D}_\mathrm{JS}(\mathcal{P}_{1},\cdots,\mathcal{P}_{n}) = H_S\Big(\sum_{i=1}^{n}\pi_{i}\mathcal{P}_{i}\Big) -\sum_{i=1}^{n} \pi_{i}H_S(\mathcal{P}_{i}),
\end{equation}
where $H_s(\mathcal{P}_{i})=\sum_{a=1}^{A}p_{ia}\mathrm{log}p_{ia}$ is the Shannon entropy of the probability distribution $\mathcal{P}_{i}$, $\pi_{i}\geq0$ is the corresponding weight for the probability distribution $\mathcal{P}_{i}$ and $\sum_{i=1}^{n}\pi_{i}=1$. In this work, we set each $\pi_{i}=\frac{1}{n}$. Since we aim to calculate the joint relevancy between features in terms of similarity measures between graph-based feature representations, we utilize the negative exponential of ${D}_\mathrm{JS}$ to calculate the similarity $I_S$ between the multiple $n$ ($n\geq2$) probability distributions, i.e.,
\begin{equation}
I_S(\mathcal{P}_{1},\cdots,\mathcal{P}_{n})=\exp\{-{D}_\mathrm{JS}(\mathcal{P}_{1},\cdots,\mathcal{P}_{n})\}.\label{Eq:SimJSD}
\end{equation}

\section{The Structural Interacting Fused Lasso}\label{s3}
In this section, we introduce the proposed approach. We commence by defining a new information theoretic measure to compute the joint relevancy between features. Moreover, we present the mathematical formulation and develop a new algorithm to solve it.
\subsection{The Proposed Information Theoretic Measure}
We propose the following information theoretic function for measuring the joint relevance of different pairwise feature combinations in relation to the target labels. For the set of $N$ features $\mathbf{f_{1}},\ldots,\mathbf{f_{i}},\ldots,\mathbf{f_{j}},\ldots,\mathbf{f_{N}}$ defined earlier and the associated discrete target feature $\mathbf{Y}$ taking the discrete values $c\in\{1,2,\ldots,C\}$, we calculate the joint relevance degree of the feature pair $\{\mathbf{f_{i}},\mathbf{f_{j}}\}$ in relation to the target feature $\mathbf{Y}$ as
\begin{equation}\label{ITC}
U_{\mathbf{f_{i}},\mathbf{f_{j}}}=\frac{I_S(\mathbf{G_{i}},\mathbf{G_{j}};\mathbf{\hat{G_{i}}})+I_S(\mathbf{G_{i}},\mathbf{G_{j}};\mathbf{\hat{G_{j}}})}
{I_{S}(\mathbf{G_{i}},\mathbf{G_{j})}},
\end{equation}
where $\mathbf{G_{i}}$ is the feature graph of each original feature $\mathbf{f_{i}}$, $\mathbf{\hat{G_{i}}}$ is the target feature graph of $\mathbf{Y}$ in terms of $\mathbf{f_{i}}$. $I_{S}(\mathbf{G_{i}},\mathbf{G_{j}})$ and $I_S(\mathbf{G_{i}},\mathbf{G_{j}};\mathbf{\hat{G_{i}}})$ are the JSD-based information theoretic similarity measures defined in Eq.(\ref{Eq:SimJSD}) for $n=2$ and $3$, respectively. The above information theoretic measure consists of two terms. The first term $I_S(\mathbf{G_{i}},\mathbf{G_{j}};\mathbf{\hat{G_{i}}})+I_S(\mathbf{G_{i}},\mathbf{G_{j}};\mathbf{\hat{G_{j}}})$ measures the relevance degrees of pairwise features $\mathbf{f_{i}}$ and $\mathbf{f_{j}}$ in relation to the target feature $\mathbf{Y}$. The second part $I_{S}(\mathbf{G_{i}},\mathbf{G_{j}})$ measures the redundancy between the feature pair $\{\mathbf{f_{i}},\mathbf{f_{j}}\}$. Therefore, the proposed structural information theoretic measure $U_{\mathbf{f_{i}},\mathbf{f_{j}}}$ is large if and only if $I_S(\mathbf{G_{i}},\mathbf{G_{j}};\mathbf{\hat{G_{i}}})+I_S(\mathbf{G_{i}},\mathbf{G_{j}};\mathbf{\hat{G_{j}}})$ is large and $I_{S}(\mathbf{G_{i}},\mathbf{G_{j}})$ is small. This indicates that the pairwise features $(\mathbf{f_{i}},\mathbf{f_{j}})$ are informative and less redundant.

Although the proposed information theoretic measure as well as that proposed by ~\cite{DBLP:journals/corr/abs-1809-02860} are both related to the JSD measure, the proposed measure differs from~\cite{DBLP:journals/corr/abs-1809-02860} in that our method focuses on the JSD measure between multiple probability distributions rather than only two probability distributions to compute the feature relevance. Therefore, the proposed information theoretic measure can compute the joint relevancy of a pair of feature combinations in relation to the target graph. By contrast, the information theoretic measure proposed by~\cite{DBLP:journals/corr/abs-1809-02860} is based upon the relevance degree of each individual feature in relation to the target feature graph, which may result in the selection of less relevant features.


Moreover, based upon the graph-based feature representations, we obtain a structural information matrix $\mathbf{U}$, where each entry $U_{i,j}\in \mathbf{U}$ corresponds to the information theoretic measure between a pair of features $\{\mathbf{f_i},\mathbf{f_j}\}$ based on Eq.(\ref{ITC}). Given the structural information matrix $\mathbf{U}$ and the $N$-dimensional feature coefficient vector $\mathbf{\beta}$, where $\beta_i$ corresponds to the coefficient of the $i$-th feature, one can locate the most discriminative feature subset by solving the optimization problem below
\begin{equation}\label{SINT}
\max f(\beta)= \max_{\beta\in \Re^{N}} \beta^{T}\mathbf{U}\beta,
\end{equation}
where $\mathbf{\beta}\geq 0$. The solution vector $\mathbf{\beta}=(\beta_{1},...,\beta_{N})^{T}$ to the above quadratic programming model is an $N$-dimensional vector. For the $i$-th positive component of $\mathbf{\beta}$, one can determine that the corresponding feature $\mathbf{f_i}$ belongs to the most discriminative feature subset, that is, feature $\mathbf{f_{i}}$ is selected if and only if $\beta_{i}>0, i\in\{1,2,...,N\}$.

\subsection{Mathematical Formulation}
Our discriminative feature selection approach is motivated by the purpose to capture structural information between pairwise features and encourage the selected features to be jointly more relevant with the target while maintaining less redundancy among them. In addition, it should simultaneously promote a sparse solution both in the coefficients and their successive differences. Therefore, we unify the minimization problem of fused lasso and Eq.(\ref{SINT}) and propose the so called fused lasso for feature selection using structural information (InFusedLasso) as
\begin{equation}\label{INT1}
\min_{\beta\in \Re^{N}}\frac{1}{2}\|\mathbf{y}-\mathbf{X}\beta\|^{2}_{2}+\lambda_{1}\|\beta\|_{1}+\lambda_{2}\|\textbf{C}\beta\|_{1}-\lambda_{3}\beta^{T}\textbf{U}\beta,
\end{equation}
where $\lambda_{1}$ and $\lambda_{2}$ are the tuning parameters for the fused lasso model, and $\lambda_{3}$ is the corresponding tuning parameter of the structural interaction matrix $\textbf{U}$. The first term in the above objective function is the error term which utilizes information from the original feature space. The second regularization term with parameter $\lambda_{1}$ encourages the sparsity of $\beta$ as in lasso and the third regularization term with parameter $\lambda_{2}$ shrinks the differences between successive features specified in matrix $\textbf{C}$ toward zero. Same as in standard fused lasso, $\textbf{C}$ is a $(N-1)\times N$ matrix with zero entries everywhere except $1$ in the diagonal and $-1$ in the superdiagonal. Moreover, the fourth term encourages the selected features to be jointly more relevant with the target while maintaining less redundancy among them. To solve the proposed model (\ref{INT1}), it is of great necessity to develop an efficient and effective algorithm to locate the optimal solutions, i.e., $\beta^{\ast}$. A feature $\mathbf{f_{i}}$ belongs to the optimal feature subset if and only if $\beta_{i}^{\ast}>0$. Accordingly, the number of optimal features can be recovered based on the number of positive components of $\beta^{\ast}$.

\subsection{Optimization Algorithm}
To effectively resolve model (\ref{INT1}), we develop an optimization algorithm based upon the split Bregman iteration approach~\cite{DBLP:journals/csda/YeX11}. We commence by reformulating the unconstrained problem (\ref{INT1}) into an equivalent constrained problem shown below
\begin{align}\label{Model3}
&\min_{\beta\in \Re^{N}}\frac{1}{2}\|\mathbf{y}-\mathbf{X}\beta\|^{2}_{2}+\lambda_{1}\|p\|_{1}+\lambda_{2}\|q\|_{1}-\lambda_{3}\beta^{T} \textbf{U}\beta\nonumber\\
&\text{s.t.}\quad p=\beta,\nonumber\\
&\quad\quad q=\textbf{C}\beta.
\end{align}
To solve the problem, we derive the split Bregman method for the proposed optimization model (\ref{Model3}) using the augmented Lagrangian method~\cite{DBLP:journals/mp/Rockafellar73}. To be specific, the corresponding Lagrangian function of (\ref{Model3}) is
\begin{equation}\label{INT3}
\widetilde{{L}}(\beta,p,q,u,v)=\frac{1}{2}\|\mathbf{y}-\mathbf{X}\beta\|^{2}_{2}-\lambda_{3}\beta^{T}\textbf{U}\beta+\lambda_{1}\|p\|_{1}+\lambda_{2}\|q\|_{1}+\langle u,\beta-p\rangle+\langle v,\textbf{C}\beta-q\rangle,
\end{equation}

where $u\in \Re^{N}$ and $v\in \Re^{N-1}$ are the dual variables corresponding to the linear constraints $p=\beta$ and $q=\textbf{C}\beta$, respectively. Here $\langle\cdot,\cdot\rangle$ denotes the standard inner product in the Euclidean space. By adding two terms $\frac{\mu_{1}}{2}\|\beta-p\|_{2}^{2}$ and $\frac{\mu_{2}}{2}\|\textbf{C}\beta-q\|_{2}^{2}$ to penalize the violation of linear constraints $p=\beta$ and $q=\textbf{C}\beta$, one can obtain the augmented Lagrangian function of (\ref{INT3}), that is,
\begin{align}
L(\beta,p,q,u,v)=\frac{1}{2}\|\mathbf{y}-\mathbf{X}\beta\|^{2}_{2}-\lambda_{3}\beta^{T}\textbf{U}\beta+\lambda_{1}\|p\|_{1}+\lambda_{2}\|q\|_{1}+ \langle u,\beta-p\rangle \nonumber \\  +\langle v,\textbf{C}\beta-q\rangle+\frac{\mu_{1}}{2}\|\beta-p\|_{2}^{2} +\frac{\mu_{2}}{2}\|\textbf{C}\beta-q\|_{2}^{2},
\end{align}
where $\mu_{1}>0$ and $\mu_{2}>0$ are the corresponding parameters.
To find a saddle point denoted as $(\beta^{*},p^{*},q^{*},u^{*},v^{*})$ for the augmented Lagrangian function ${\L}(\beta,p,q,u,v)$, the following inequalities hold
\begin{equation}\label{INT5}
L(\beta^{*},p^{*},q^{*},u,v)\leq L(\beta^{*},p^{*},q^{*},u^{*},v^{*})\leq L(\beta,p,q,u^{*},v^{*}),
\end{equation}
for all $\beta,p,q,u$ and $v$. It is clear that $\beta^{*}$ is an optimal solution to (\ref{INT1}) if and only if $\beta^{*}$, $p^{*}$, $q^{*}$, $u^{*}$ and $v^{*}$ solves this saddle point problem for some $p^{*}$, $q^{*}$, $u^{*}$ and $v^{*}$~\cite{Rockafellar1997}.

We solve the above saddle point problem using an iterative algorithm by alternating between the primal and the dual optimization shown below
\begin{equation}\label{Model5}
\left\{
\begin{array}{ll}
\text{Primal:} (\beta^{k+1},p^{k+1},q^{k+1})=\mathop{\arg\min}\limits_{\beta,p,q}L(\beta,p,q,u^{*},v^{*})\nonumber\\
\text{Dual:}  u^{k+1}=u^{k}+\delta_{1}(\beta^{k+1}-p^{k+1})
\nonumber\\
\quad\quad v^{k+1}=v^{k}+\delta_{2}(\textbf{C}\beta^{k+1}-q^{k+1}),
\end{array}\right.
\end{equation}
where the first step updates the primal variables based upon the current estimation of $u^{k}$ and $v^{k}$, followed by the second step which updates the dual variables based upon the current estimates of the primal variables. Because the augmented Lagrangian function is linear in both $u$ and $v$, updating the dual variables is comparatively simple and we adopt a gradient ascent method with step size $\delta_{1}$ and $\delta_{2}$. Therefore, the efficiency of the above optimization algorithm depends upon whether the primal problem can be resolved quickly. To facilitate better illustration, denote
\begin{equation}
V(\beta)=\frac{1}{2}\|\mathbf{y}-\mathbf{X}\beta\|^{2}_{2}-\lambda_{3}\beta^{T}\textbf{U}\beta
+\langle u^{k},\beta-p^{k}\rangle+\langle v^{k},\textbf{C}\beta-q^{k}\rangle+\frac{\mu_{1}}{2}\|\beta-p^{k}\|_{2}^{2}+\frac{\mu_{2}}{2}\|\textbf{C}\beta-q^{k}\|_{2}^{2}.
\end{equation}

Because the objective function on minimizing $\beta$, i.e.,$V(\beta)$ is differentiable, we can resolve the primal problem by alternatively minimizing $\beta$, $p$, and $q$ as follows.
\begin{equation}\label{INT5}
\left\{
\begin{array}{ll}
\beta^{k+1}=\mathop{\arg\min}\limits_{\beta}V(\beta^{k})\nonumber\\
p^{k+1}=\mathop{\arg\min}\limits_{p}\lambda_{1}\|p\|_{1}+\langle u^{k},\beta^{k+1}-p\rangle+\frac{\mu_{1}\|\beta^{k+1}-p\|_{2}^{2}}{2}\nonumber\\
q^{k+1}=\mathop{\arg\min}\limits_{q}\lambda_{2}\|q\|_{1}+\langle v^{k},\textbf{C}\beta^{k+1}-q\rangle+\frac{\mu_{2}\|\textbf{C}\beta^{k+1}-q\|_{2}^{2}}{2}.
\end{array}\right.
\end{equation}
Furthermore, since the objective function $V(\beta)$ on minimizing $\beta$ is quadratic and differentiable, we can obtain the optimal solution of $\beta$ by setting $\frac{\partial(V(\beta))}{\partial\beta}=0$, that is,
\begin{equation}\label{INT8}
\mathbf{X}^{T}\mathbf{X}\beta-2\lambda_{3} \textbf{U}\beta+\mu_{1}\textbf{I}\beta+\mu_{2}\textbf{C}^{T}\textbf{C}\beta-\mathbf{X}^{T}y
-\mu_{1}p^{k}+\mu_{1}\mu_{1}^{-1}u^{k}-\mu_{2}\textbf{C}^{T}q^{k}+\mu_{2}\textbf{C}^{T}\mu_{2}^{-1}v^{k}=0,
\end{equation}
i.e.,the optimal solution are obtained by solving a set of linear equations as follows
\begin{equation}\label{INT9}
\textbf{D}\beta^{k+1}=\mathbf{X}^{T}y+\mu_{1}(p^{k}-\mu_{1}^{-1}u^{k})+\mu_{2}\textbf{C}^{T}(q^{k}-\mu_{2}^{-1}v^{k}).
\end{equation}
Because matrix $\textbf{D}=\mathbf{X}^{T}\mathbf{X}-2\lambda_{3}\textbf{U}+\mu_{1}\textbf{I}+\mu_{2}\textbf{C}^{T}\textbf{C}$ is a $N\times N$ matrix, which is independent of the optimization variables. For small $N$, we can invert $\textbf{D}$ and store $\textbf{D}^{-1}$ in the memory, such that the linear equations are resolved with minimum cost. That is, $\beta^{k+1}=\textbf{D}^{-1}(\mathbf{X}^{T}y+\mu_{1}(p^{k}-\mu_{1}^{-1}u^{k})+\mu_{2}\textbf{C}^{T}(q^{k}-\mu_{2}^{-1}v^{k}))$. However, for large $N$, we need to numerically solve the linear equations at each iteration by means of the conjugate gradient algorithm.

Overall, we develop Algorithm~\ref{alg:algorithm} for locating optimal solutions to the proposed feature selection problem, where minimization of $p$ and $q$ are achieved efficiently through soft thresholding. Note that $\Gamma_{\lambda}(\omega)=[t_{\lambda}(\omega_{1},t_{\lambda}(\omega_{1},...,...]^{T}$, with $t_{\lambda}(\omega_{i})=\text{sgn}(\omega_{i})\max\{0,|\omega_{i}|-\lambda\}$. Furthermore, we can follow the same idea as in ~\cite{DBLP:journals/csda/YeX11} and obtain the convergence proof of the algorithm.
\begin{algorithm}[tb]
\caption{The proposed iterative optimization algorithm}
\label{alg:algorithm}
\textbf{Input}: $\mathbf{X},\mathbf{y},\beta^{0},p^{0},q^{0}, u^{0}$ and $v^{0}$.\\
\textbf{Output}: $\beta^{*}$
\begin{algorithmic}[1] 
\WHILE{not converged}
\STATE Update $\beta^{k+1}$ according to the solution to Eq.(\ref{INT9}).
\STATE Update $p^{k+1}$ with $p^{k+1}=\Gamma_{\mu_{1}^{-1}\lambda_{1}}(\beta^{k+1}+\mu_{1}^{-1}u^{k})$.
\STATE Update $q^{k+1}$ with $q^{k+1}=\Gamma_{\mu_{2}^{-1}\lambda_{2}}(\textbf{C}\beta^{k+1}+\mu_{2}^{-1}v^{k})$.
\STATE Update $u^{k+1}$ with $u^{k+1}=u^{k}+\delta_{1}(\beta^{k+1}-p^{k+1})$.
\STATE Update $v^{k+1}$ with $v^{k+1}=v^{k}+\delta_{2}(\textbf{C}\beta^{k+1}-q^{k+1})$.
\ENDWHILE
\STATE \textbf{return} solution
\end{algorithmic}
\end{algorithm}

\section{Experiments}\label{s4}
In this section, we conduct several experiments on standard machine learning datasets to verify the performance of the proposed fused lasso feature selection method (InFusedLasso). These datasets are abstracted from Biomedical, Speech, Text and Computer Vision databases. Details of these datasets are presented in Table~\ref{table1}.
\subsection{Experimental Settings and Results}
\begin{table}
 \vspace{-5pt}
\centering
\footnotesize
\caption{Statistics of experimental datasets}
 \vspace{-7pt}
\begin{tabular}{llrrr}
\toprule
~Type~ &~Name ~  &~\#Feature~     &~\#Sample~  &~\#Class~ \\
\midrule
~Image~ &~USPS~      &~256~      &~9298~     &~10~  \\

~ ~     &~Pie~       &~1024~     &~11554~    &~68~ \\

~ ~     &~YaleB~     &~1024~     &~2414~     &~38~ \\

~Biomedical~ &~Lymphoma~  &~4026~     &~96~       &~9~  \\

~ ~        &~Leukemia~  &~7129~     &~73~       &~2~  \\

~Text~ &~RELATHE~   &~4322~     &~1427~     &~2~ \\

~ ~ & ~BASEHOCK~  &~4862~     &~1993~     &~2~\\

~Speech~ &~Isolet1~   &~617~      &~1560~     &~26~\\
\bottomrule
\end{tabular}
\label{table1}
 \vspace{-15pt}
\end{table}

\begin{figure*}
 \centering
   \vspace{-15pt}
    \subfigure[For USPS]{\includegraphics[width=0.2425\linewidth]{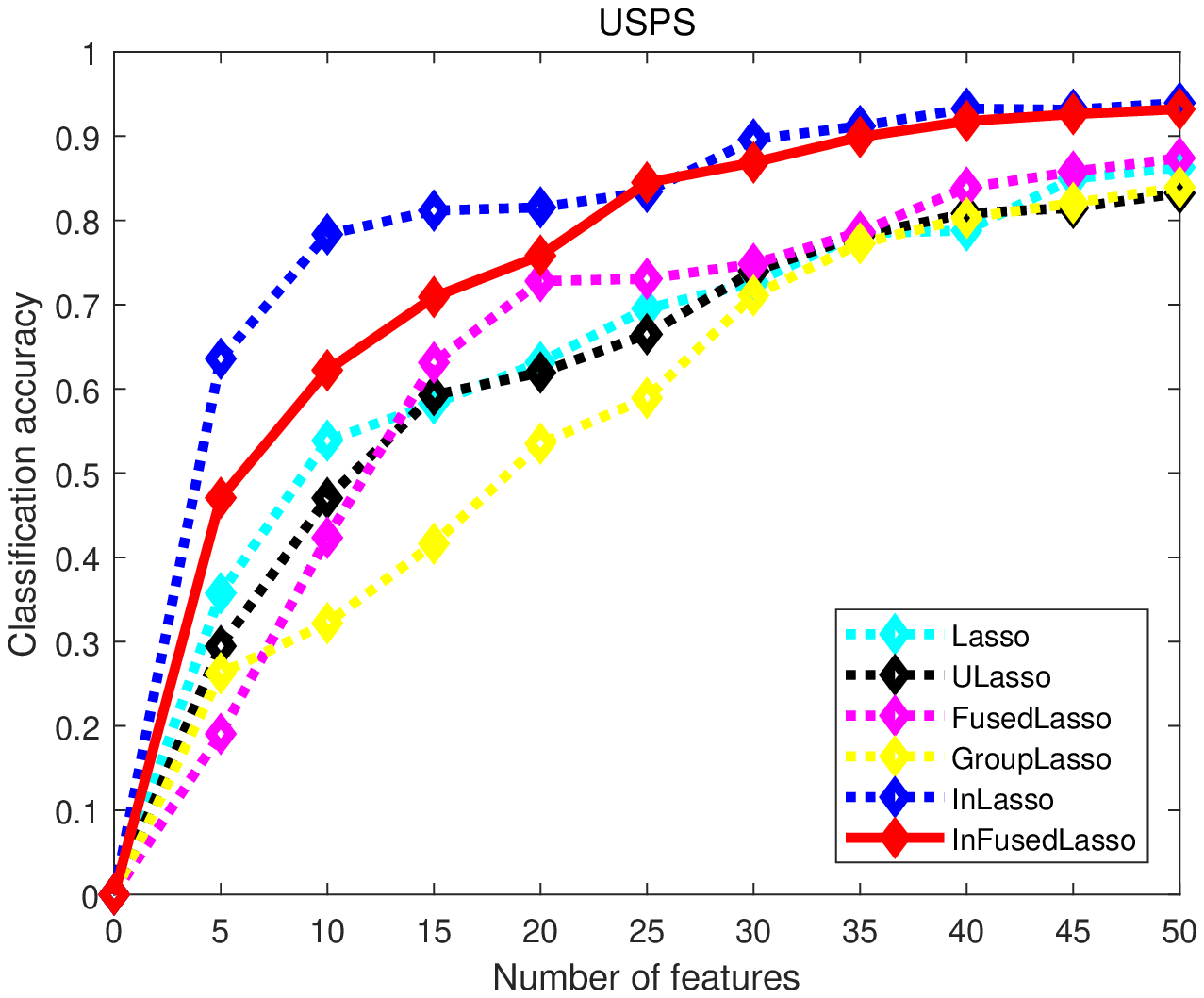}}
    \subfigure[For Pie]{\includegraphics[width=0.2425\linewidth]{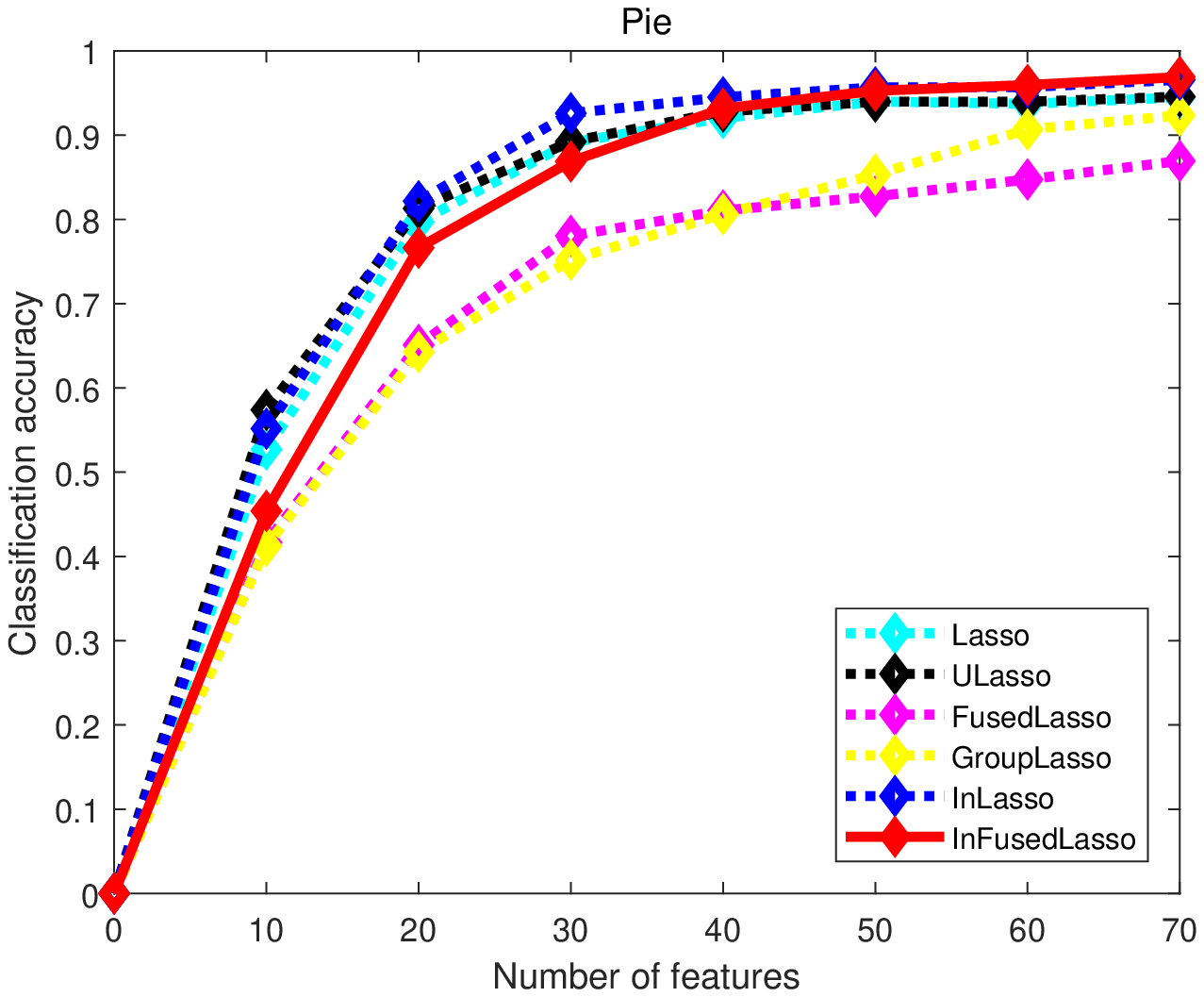}}
    \subfigure[For YaleB]{\includegraphics[width=0.2425\linewidth]{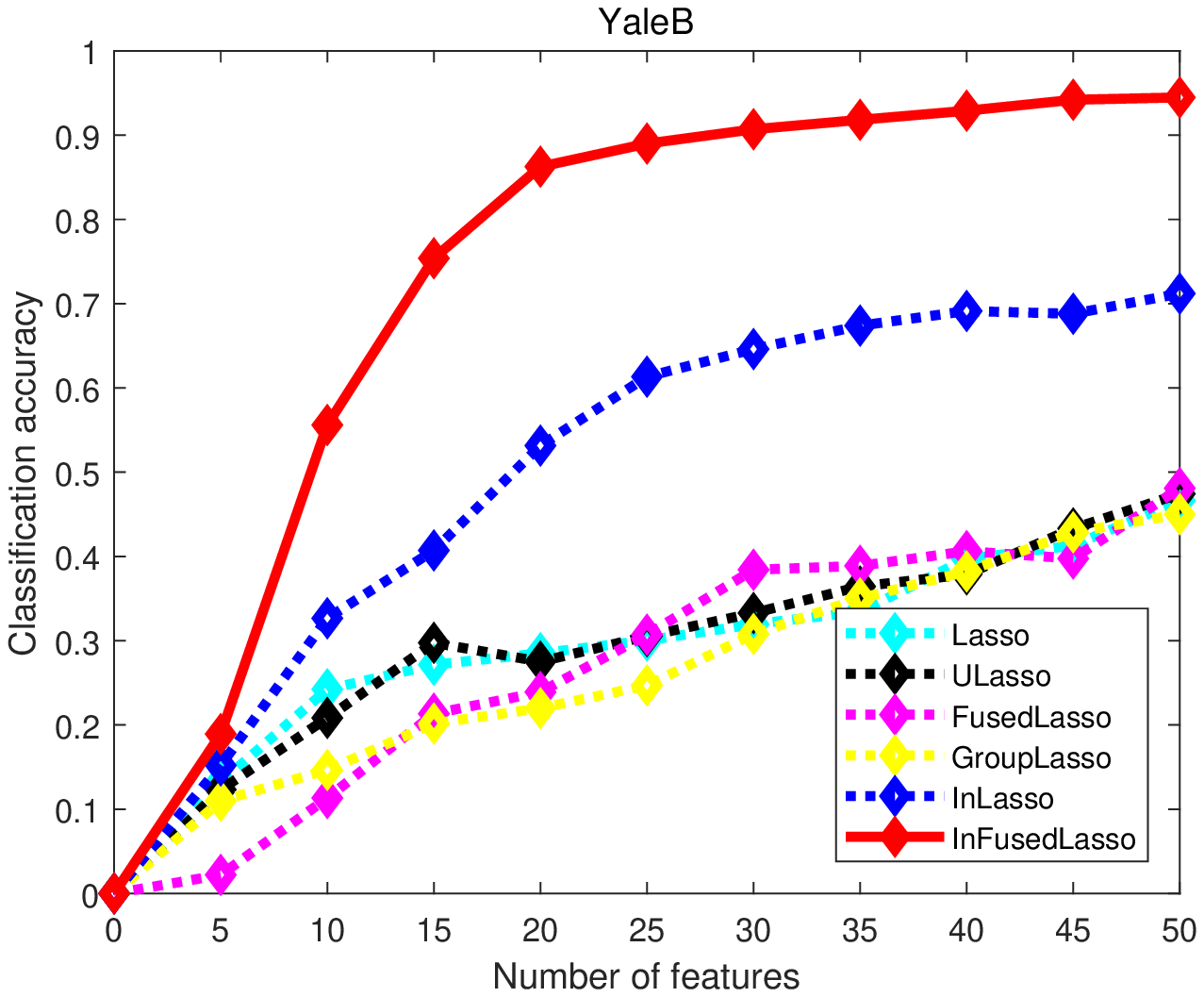}}
    \subfigure[For Lymphoma]{\includegraphics[width=0.2425\linewidth]{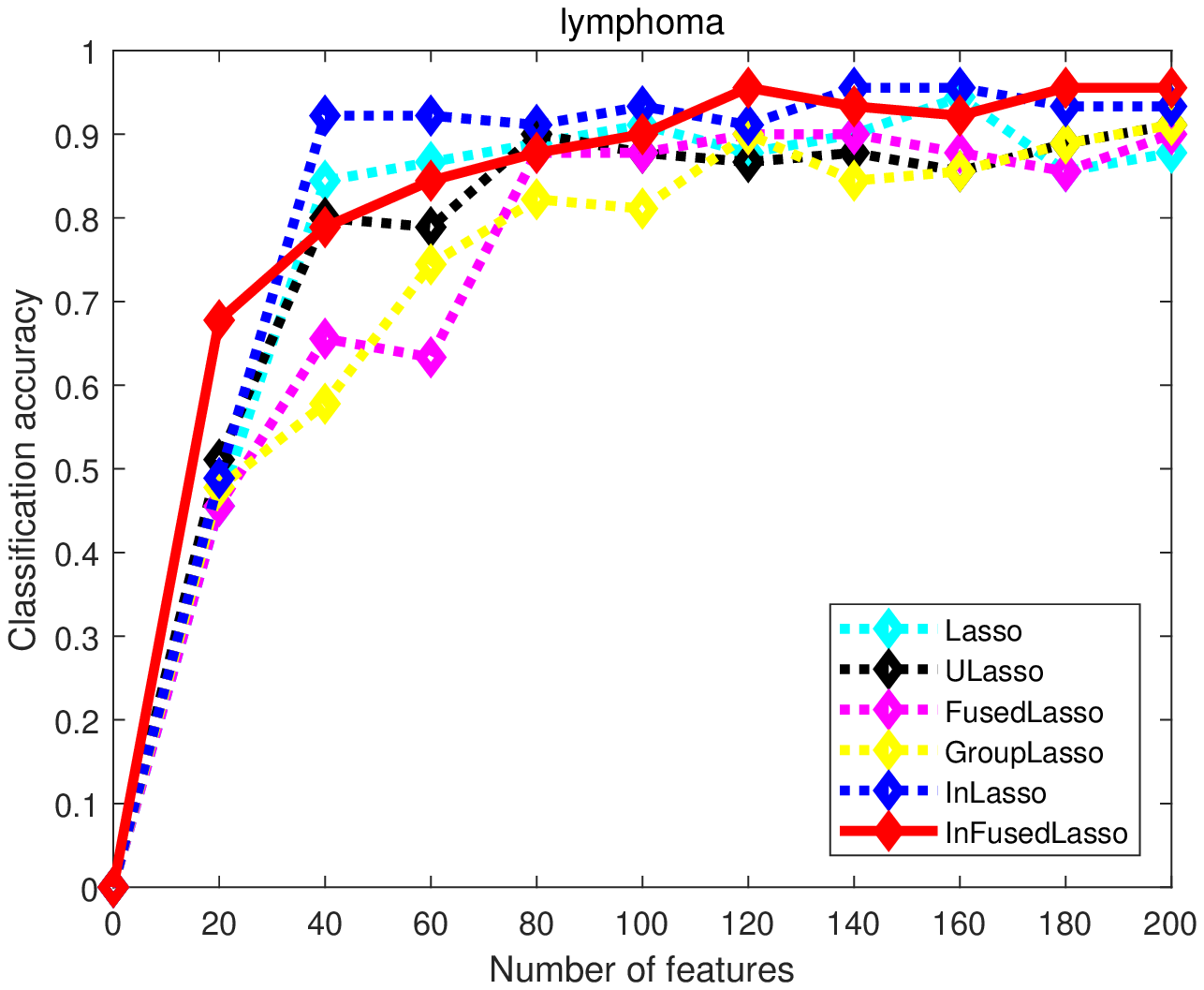}}
    \subfigure[For Leukemia]{\includegraphics[width=0.2425\linewidth]{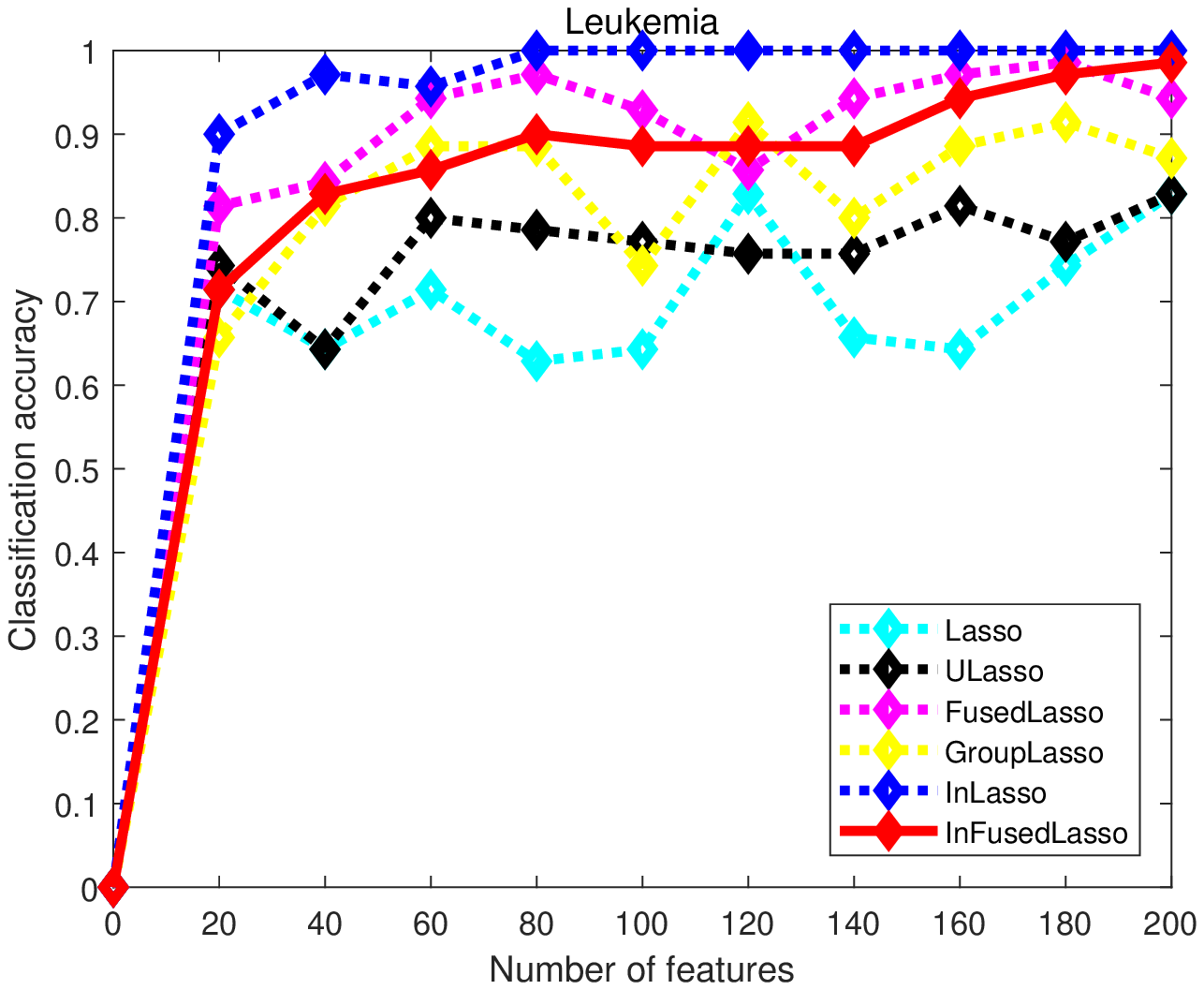}}
    \subfigure[For RELATHE]{\includegraphics[width=0.2425\linewidth]{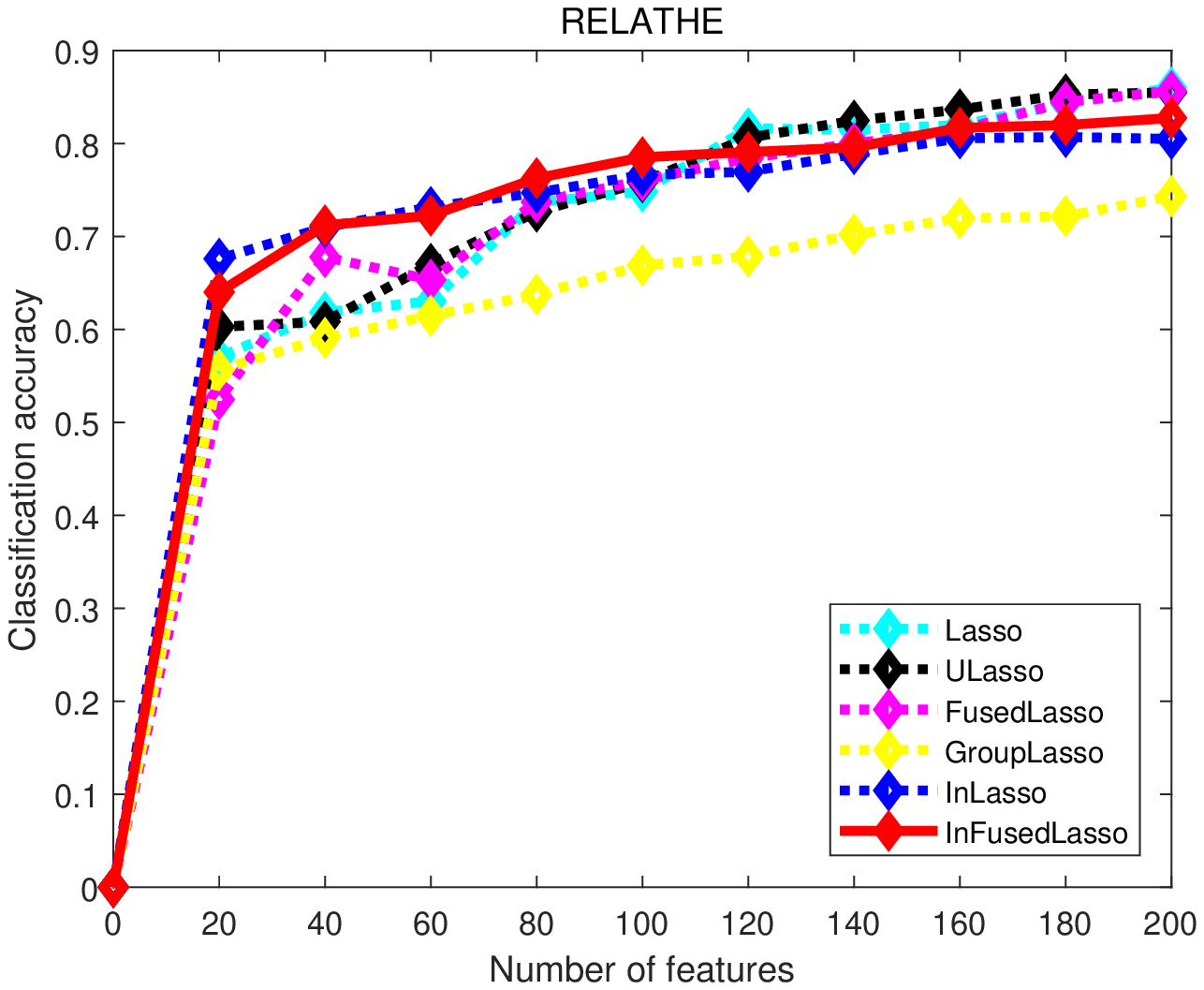}}
    \subfigure[For BASEHOCK]{\includegraphics[width=0.2425\linewidth]{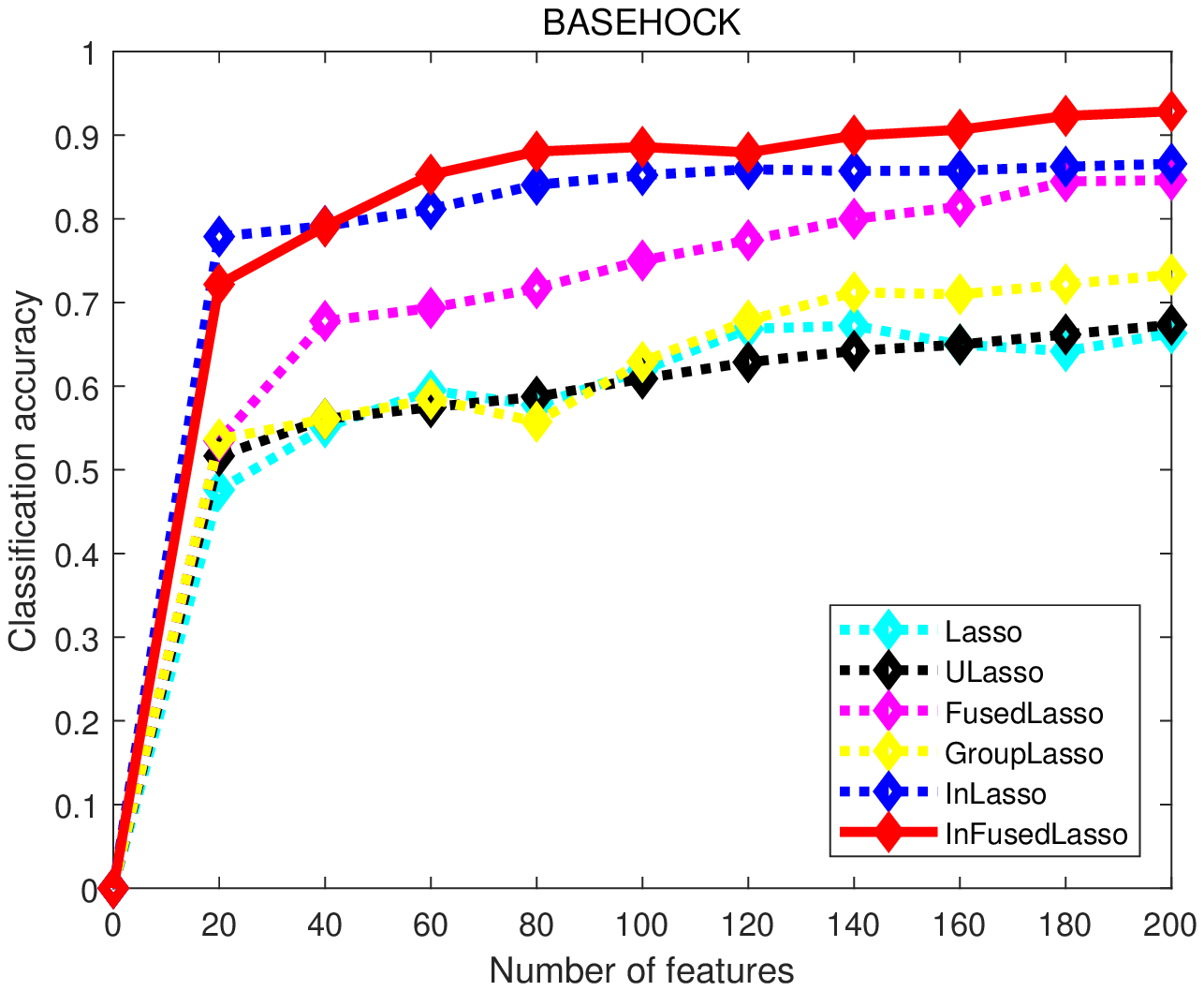}}
    \subfigure[For Isolet1]{\includegraphics[width=0.2425\linewidth]{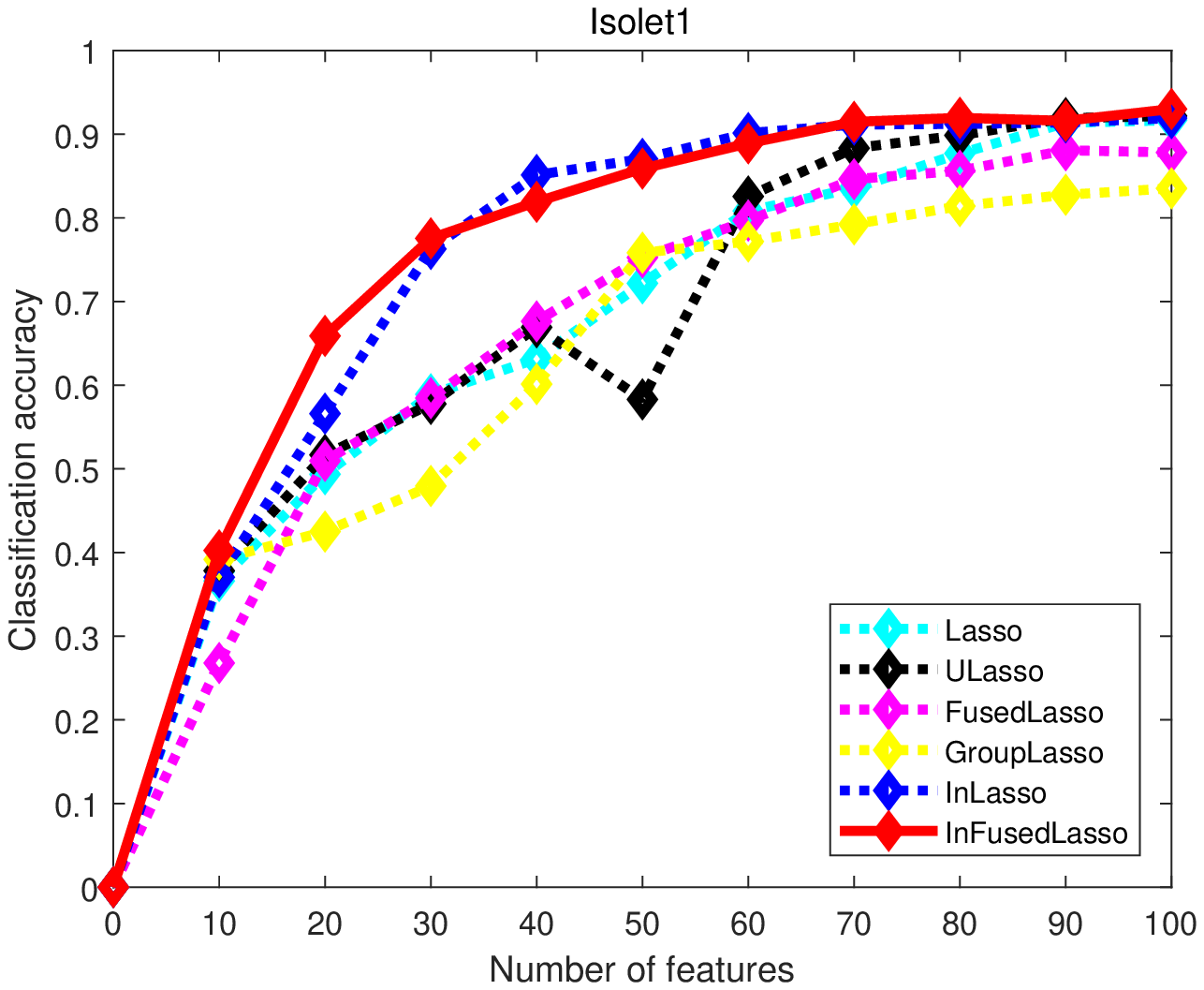}}
    \vspace{-10pt}
    \caption{Accuracy versus the number of selected features.}\label{stabilityEntropy}
    \vspace{-10pt}
\end{figure*}

\begin{table*}
\centering{
 \tiny
\caption{The best results of all methods and the corresponding number of selected features.}\label{Acc}
 \vspace{-0pt}
\begin{tabular}{rrrrrrrrr}
\toprule
~Dataset~    &~USPS~          &~Pie~           & ~YaleB~         & ~Lymphoma~       & ~Leukemia~      & ~RELATHE~        & ~BASEHOCK~      & ~Isolet1~ \\ \hline

~Lasso~      &~$86.30 (50)$~  &~$94.48 (70)$~  & ~$46.64 (50)$~  & ~$91.11 (100)$~  & ~$82.86 (200)$~ & ~$\textbf{86.00} (200)$~  & ~$67.22 (140)$~ & ~$91.67 (100)$~\\

~ULasso~    &~$83.25 (50)$~  &~$94.57 (70)$~  & ~$47.43 (50)$~  & ~$94.44 (160)$~  & ~$82.86 (200)$~ & ~$85.49 (200)$~  & ~$67.33 (200)$~ & ~$92.18 (100)$~\\

~FusedLasso~    &~$87.40 (50)$~  &~$86.94 (70)$~  & ~$48.09 (50)$~  & ~$90.00 (120)$~  & ~$94.29 (140)$~ & ~$85.62 (200)$~  & ~$84.62 (200)$~ & ~$88.08 (~~90)$~\\

~GroupLasso~ &~$83.93 (50)$~  &~$92.35 (70)$~  & ~$45.02 (50)$~  & ~$91.11 (200)$~  & ~$91.43 (180)$~ & ~$74.33 (200)$~  & ~$73.33 (200)$~ & ~$83.53 (100)$~\\

~InLasso~    &~$\textbf{93.94} (50)$~  &~$96.58 (70)$~  & ~$71.20 (50)$~  & ~$95.55 (140)$~  & ~$\textbf{100.00} (~~80)$~ & ~$80.70 (180)$~  & ~$86.58 (180)$~ & ~$91.92 (100)$~\\
\midrule
~\textbf{InFusedLasso}~  &~$93.67 (50)$~  &~$\textbf{96.90} (70)$~  & ~$\textbf{94.87} (50)$~  & ~$\textbf{95.55} (120)$~  & ~$98.57 (200)$~ & ~$84.50 (200)$~  & ~$\textbf{92.85} (200)$~ & ~$\textbf{93.01} (100)$~\\
\bottomrule
\end{tabular}

\caption{InFusedLasso versus InElasticNet.}\label{table3}
 \vspace{-0pt}
\begin{tabular}{rrrrrrrrr}
\toprule
~Dataset~    &~USPS~          &~Pie~           & ~YaleB~         & ~Lymphoma~       & ~Leukemia~      & ~RELATHE~        & ~BASEHOCK~      & ~Isolet1~ \\ \hline

~InElasticNet~   &~$\textbf{94.10} (50)$~  &~$96.82 (70)$~  & ~$94.62 (50)$~  & ~$95.55(160) $~  & ~$ \textbf{100.00}(80)$~ & ~$83.02 (200)$~  & ~$92.75 (200)$~ & ~$ 92.03 (100)$~\\

~\textbf{InFusedLasso}~  &~$93.67 (50)$~  &~$\textbf{96.90} (70)$~  & ~$\textbf{94.87} (50)$~  & ~$\textbf{95.55} (120)$~  & ~$98.57 (200)$~ & ~$\textbf{84.50} (200)$~  & ~$\textbf{92.85}(200)$~ & ~$\textbf{93.01} (100)$~\\
\bottomrule
\end{tabular}
} \vspace{-0pt}
\end{table*}

To validate the effectiveness of the proposed InFusedLasso method, we compare the classification accuracies on different datasets based on the features selected from the proposed method. We also compare the proposed method to several existing state-of-the-art lasso-type feature selection methods. These methods for comparisons include Lasso~\cite{Lasso}, Fused Lasso~\cite{FusedLasso}, Group Lasso~\cite{DBLP:journals/bmcbi/MaSH07}, ULasso~\cite{DBLP:conf/aaai/ChenDLX13}, and InLasso~\cite{DBLP:journals/prl/ZhangTBXH17}. For the experiments, we utilize a 10-fold cross-validation approach associated with a C-SVM classifier based on the Linear kernel to evaluate the classification accuracies. We use nine folders for training and one folder for testing. We repeat the whole experiment 10 times, and the performance of various feature selection methods is evaluated in terms of the mean classification accuracies versus different number of selected features. The results are shown in Figure~\ref{stabilityEntropy}. In addition, the best mean classification accuracies of different methods associated with the number of selected features are reported in Table~\ref{Acc}.


Figure~\ref{stabilityEntropy} exhibits the advantage of the proposed InFusedLasso method. When the number of selected features reaches a certain number, the proposed approach can outperform the alternative methods on the Pie, YaleB, Lymphoma, BASEHOCK and Isolet1 datasets. Moreover, Table~\ref{Acc} confirms that the proposed approach can achieve the best classification performance on the Pie, YaleB, Lymphoma, BASEHOCK, and Isolet1 datasets. On the other hand, for the USPS, Leukemia, and RELATHE datasets, although the proposed method cannot achieve the best classification accuracy, the proposed method is still competitive to the alternatives. The experimental result indicates that the proposed InFusedLasso method can better learn the characteristics and interaction information residing on the features. This is because only the proposed method can encapsulate the structure correlated information between feature samples through the structure-based feature graph representation. To take our study one step further, we also compare the proposed InFusedLasso method to the Interacted ElasticNet method (InElaNet)~\cite{DBLP:journals/corr/abs-1809-02860}, since this method can also encapsulate the structure correlated information. Table~\ref{table3} indicates that the proposed method can outperform the InElaNet method on most of the datasets. The reason is that the required feature graph structures of the InElaNet method is computed based on the Eucliden distance. As we have stated earlier, the distance with large value may dominant the characteristics of the feature graph and influence the effectiveness. By contrast, the proposed InFusedLasso method employs a new kernel-based graph modeling procedure to establish feature graphs and proposes a new information theoretic criterion to measure the joint relevancy of pairwise feature combinations in relation to the discrete target. As a result, the proposed InFusedLasso method overcomes the shortcomings of the InElaNet method.

Overall, the experimental results verify that the proposed approach can identify more informative feature subsets than state-of-the-art feature selection methods.

\subsection{Convergence Evaluation}
\begin{figure}
 \centering
  \vspace{-10pt}
    \subfigure[For USPS]{\includegraphics[width=0.3\linewidth]{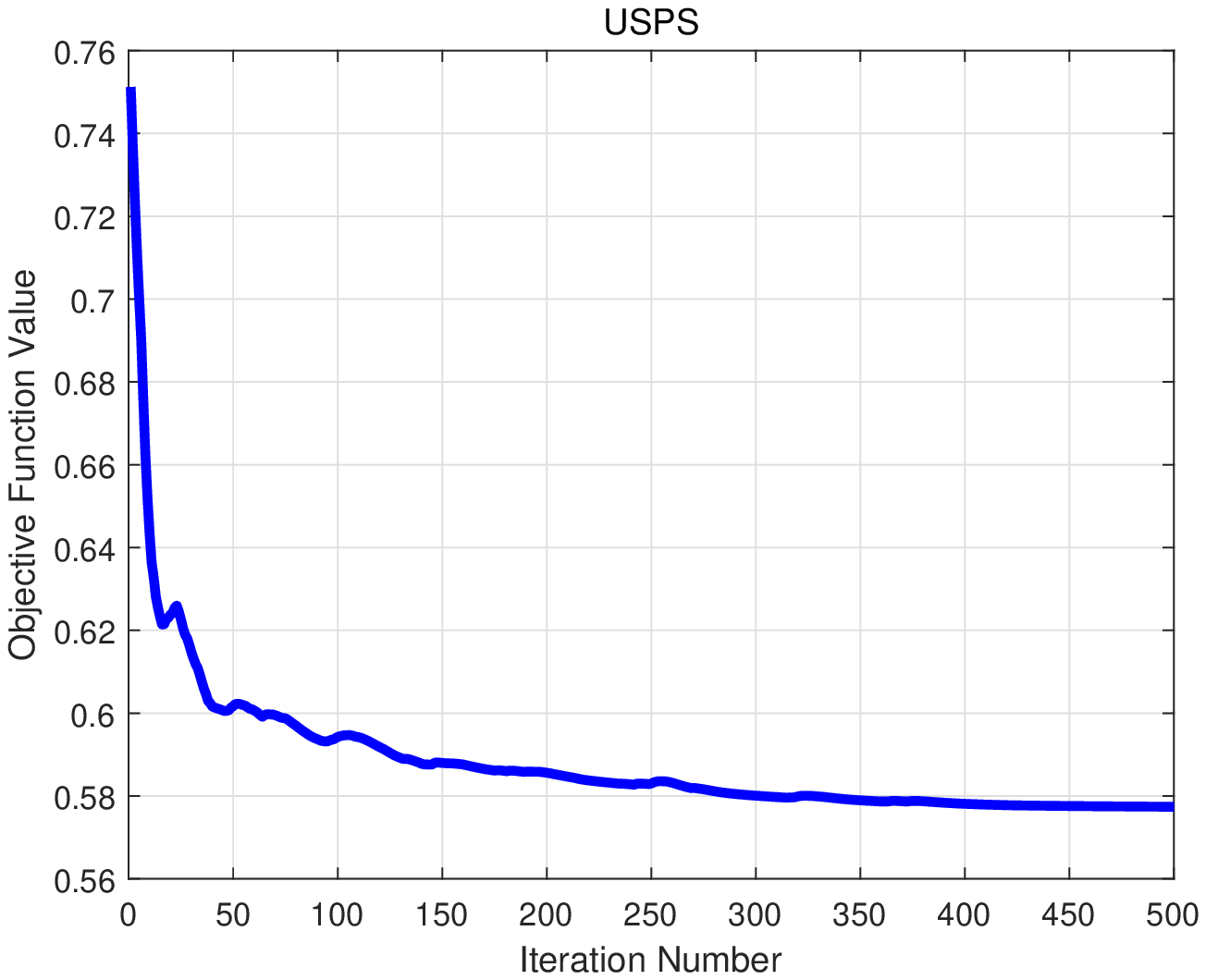}}
    \subfigure[For YaleB]{\includegraphics[width=0.3\linewidth]{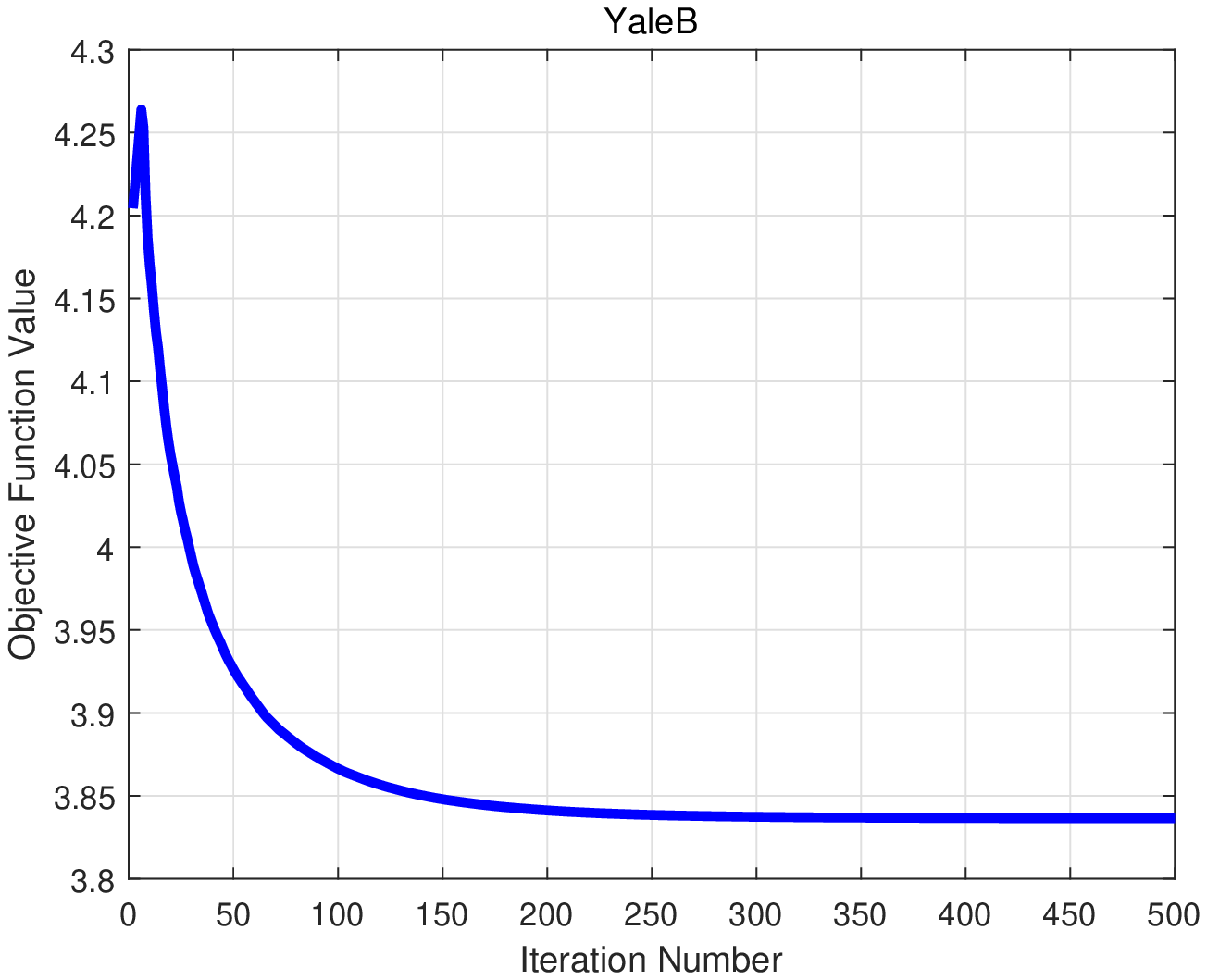}}
    \vspace{-10pt}
    \caption{Convergence curve for the optimization algorithm.}\label{Convergence}
    \vspace{-20pt}
\end{figure}
In this subsection, we experimentally evaluate the convergence properties of the proposed optimization algorithm. Due to the limited space of the manuscript, we only display the convergence curves on two datasets, i.e., USPS and YaleB. Note that, we can observe similar results on the remaining datasets. Specifically, the variations of the objective function values at each iteration are reported in Figure~\ref{Convergence}. Figure~\ref{Convergence} indicates that the proposed optimization algorithm converges as the iteration number within about 150 iterations, which ensures the efficiency and effectiveness of the proposed feature selection approach.

\section{Conclusion}\label{s6}
In this paper, we have developed a new fused lasso feature selection with structural information. The proposed approach incorporates structural correlation information between pairwise samples into the feature selection process, and has the ability to maximize joint relevance of pairwise feature combinations in relation to the target and minimize redundancy of selected features. In addition, the proposed method can promote sparsity in the features and their successive neighbors. An effective iterative algorithm is proposed to solve the proposed feature subset selection problem based on the split Bregman method. Experiments demonstrate the effectiveness of the proposed feature selection approach.
\small{
\bibliographystyle{named}
\bibliography{ijcai19}
}

%
%
%
%
%

\end{document}